\documentclass{article}

\PassOptionsToPackage{numbers}{natbib}
\usepackage[preprint]{neurips_data_2024}

\usepackage[utf8]{inputenc} 
\usepackage[T1]{fontenc}    
\usepackage{hyperref}       
\usepackage{url}            
\usepackage{booktabs}       
\usepackage{amsfonts}       
\usepackage{nicefrac}       
\usepackage{microtype}      
\usepackage{xcolor}         
\usepackage{graphicx}
\usepackage{amsmath}
\usepackage{xspace}
\usepackage{multirow}
\usepackage{array}  
\usepackage{pdfpages}
\usepackage{tikz}
\usetikzlibrary{matrix}

\newcommand{\data}{\textsc{MR-GSM8K}\xspace}
\title{\data: A Meta-Reasoning Benchmark  \\for Large Language Model Evaluation}

\author{%
  Zhongshen Zeng \\
  Chinese University of HongKong\\
  \texttt{zszeng23@cse.cuhk.edu.hk} \\
  \And
  Pengguang Chen \\
Chinese University of Hong Kong \\
\texttt{pgchen@cse.cuhk.edu.hk} \\\AND
Shu Liu \\
Smartmore Co.Ltd \\
\texttt{sliu@smartmore.com} \And
Haiyun Jiang  \\
Tencent AI Lab \\
\texttt{haiyunjiang@tencent.com} \And
Jiaya Jia  \\
Chinese University of Hong Kong \\
\texttt{leojia@cse.cuhk.edu.hk}
}

\begin{document}

\maketitle

\begin{abstract}
In this work, we introduce a novel evaluation paradigm for Large Language Models (LLMs) that compels them to transition from a traditional question-answering role, akin to a student, to a solution-scoring role, akin to a teacher. This paradigm, focusing on "reasoning about reasoning," hence termed meta-reasoning, shifts the emphasis from result-oriented assessments, which often neglect the reasoning process, to a more comprehensive evaluation that effectively distinguishes between the cognitive capabilities of different models. By applying this paradigm in the GSM8K dataset, we have developed the MR-GSM8K benchmark. Our extensive analysis includes several state-of-the-art models from both open-source and commercial domains, uncovering fundamental deficiencies in their training and evaluation methodologies. Notably, while models like Deepseek-v2 and Claude3-Sonnet closely competed with GPT-4 in GSM8K, their performance disparities expanded dramatically in MR-GSM8K, with differences widening to over 20 absolute points, underscoring the significant challenge posed by our meta-reasoning approach.
\end{abstract}

\section{Introduction} \label{sec:intro}

\begin{figure*}[ht]
\centering
\includegraphics[width=\textwidth]{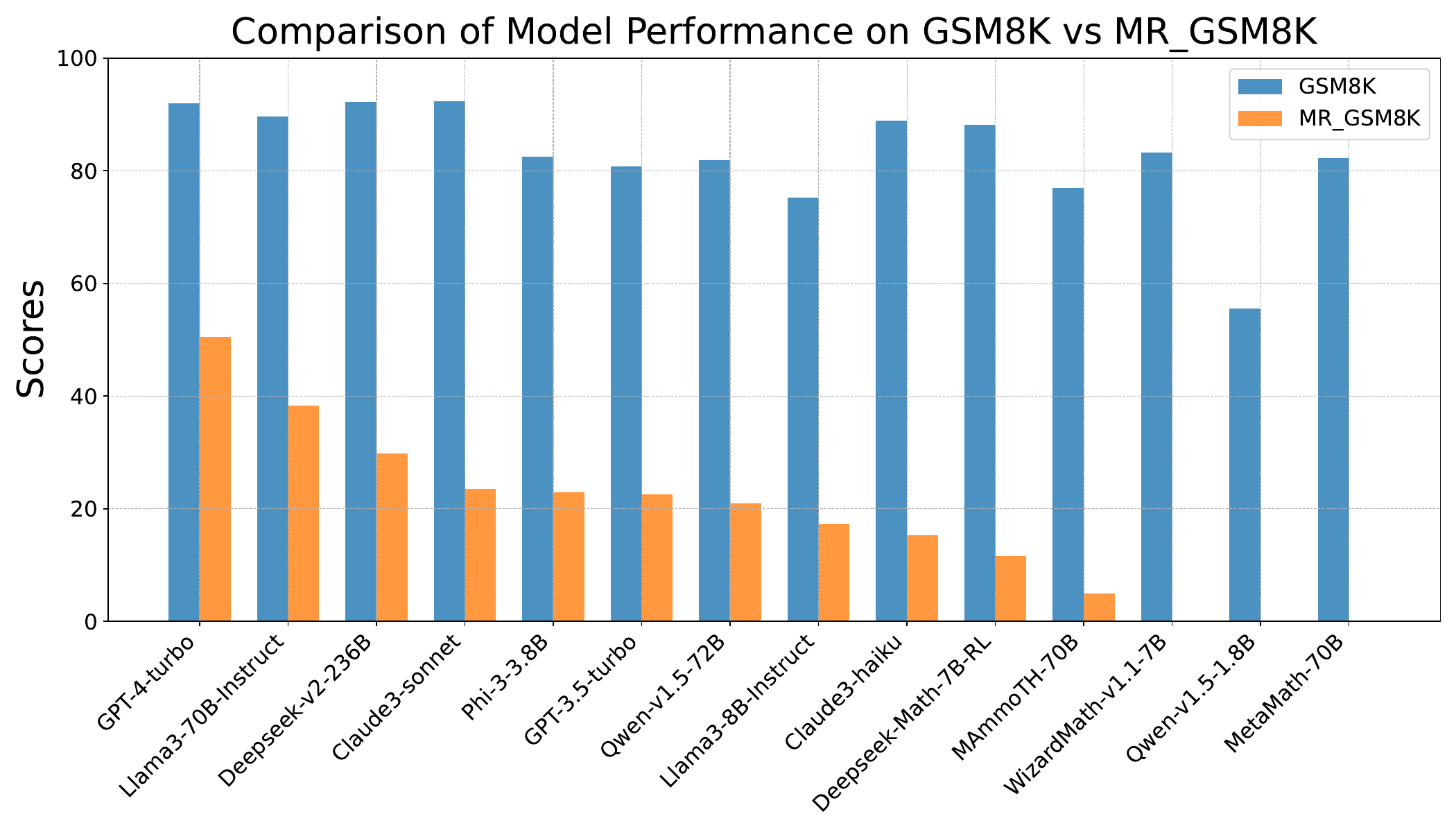}
\vspace{-0.8cm}
\caption{Model performances on GSM8K measured in accuracy versus performances on \data measured in MR-Score. Note that some models score near zero in our benchmark, highlighting the limitations of current training and evaluation paradigms. }
\label{fig:gsm_vs_mrgsm}
\end{figure*}

Pretrained on trillions of tokens and equipped with billions of parameters, today’s large language models \citep{OpenAI2023GPT4TR, ANTHROPIC2023CLAUDE, touvron2023llama} are capable of generating coherent texts and achieving super-human performances in many tasks \citep{Bubeck2023SparksOA, hendrycks2021measuring}. With the aim of differentiating cognitive abilities among models, math questions are often chosen as a proxy evaluation task. However, despite the complexity and diversity of these math problems, recent state-of-the-art LLMs \citep{OpenAI2023GPT4TR, yu2023metamath, Gou2023ToRAAT} have achieved accuracy rates exceeding 80\% \citep{luo2023wizardmath} on multi-step math reasoning datasets like GSM8K \citep{cobbe2021training}.

Upon detailed examination of the design principles and objectives of current math datasets, we identified several key shortcomings. Firstly, the majority of these datasets focus on result-oriented metrics, scoring accuracy based solely on the final answer, without considering the underlying reasoning process. With the emergence of the chain of thought methodology \citep{Wei2022ChainOT} and its derivative techniques \citep{Chen2022ProgramOT, Yao2023TreeOT} as the de facto standard for evaluating reasoning processes, we argue that the result-driven evaluation method may be insufficient for a comprehensive assessment of intended cognitive and reasoning capabilities. Secondly, a recent study \citep{PasterHungarianTesting} suggests that some LLMs, which achieved state-of-the-art performances in GSM8K and MATH \citep{hendrycks2021measuring} benchmarks, demonstrate unexpectedly low performance when facing newly released Hungarian high school exams. This raises concerns about data contamination and potential overfitting to the benchmarks, and it challenges the efficacy of these benchmarks in differentiating model capabilities.

In response to these identified limitations, we introduced a novel paradigm that shifts the role of the evaluated model from a question-answering student to a solution-scoring teacher. Specifically, instead of delivering potential solutions to given questions, which may be subject to data contamination issues, the evaluated models are now presented with question-solution pairs and tasked with determining solution correctness, identifying potential first error steps, and providing reasons for errors. This paradigm shift challenges the evaluated models to engage in meta-reasoning about different reasoning processes, a concept we term "meta-reasoning" in this paper.

Following this design principle, we have developed a new benchmark named \textbf{Meta-Reasoning-GSM8k} (\textbf{MR-GSM8k}) and proposed a novel metric called MR-Score. Our benchmark, characterized by instances manually labeled by experts and rigorously reviewed, serves as a robust tool for both qualitative and quantitative assessments of language models. Our findings indicate that most state-of-the-art models demonstrate a significant performance decline in this more nuanced assessment. As demonstrated in Figure-\ref{fig:gsm_vs_mrgsm}, although state-of-the-art models exhibit comparable performance in GSM8K, there is considerable variance in their effectiveness on our benchmark, with discrepancies up to more than tenfold.

We argue that our evaluation paradigm not only introduces a metric that more effectively differentiates based on the reasoning process over mere computational outcomes, but it also exposes fundamental deficiencies within current evaluation and training methodologies. First, as detailed in Section-\ref{sec:experiments}, our experiments revealed that specialized math models struggle to generalize their reasoning abilities to our new paradigm, regardless of whether they are directed by specific instructions or through few-shot in-context learning. Second, we observed that contrary to the common belief that larger models inherently perform better in reasoning tasks, smaller chat models such as Phi-3-Mini \cite{DBLP:journals/corr/abs-2306-11644} can surpass those twenty times their size. This underscores the significance of data quality and techniques like knowledge distillation in enhancing model performance. Third, our empirical findings indicate that current models still engage in superficial mathematical reasoning, exhibiting flaws such as a lack of ontological understanding \cite{Toh2023VerityMathAM}, susceptibility to the reversal curse \cite{DBLP:journals/corr/abs-2309-12288}, and inadequate global comprehension of solution spaces.   

In conclusion, our paper significantly contributes to the field in the following ways:
\begin{itemize}
\item We introduce a novel evaluation principle, the accompanying open-source benchmark MR-GSM8k, and the metric MR-Score.
\item We demonstrate the effective transformation of an existing benchmark (e.g., GSM8K) and how such modification can lead to robust evaluation against potential overfitting and data contamination.

\item We conduct comprehensive experiments on an array of state-of-the-art models using the MR-GSM8k benchmark, highlighting critical shortcomings in current training and evaluation paradigms.

\item Through analysis of cognitive levels and examination of holistic coverage on the solution space, we emphasize the need for benchmarks that go beyond surface-level evaluations, fostering more sophisticated and nuanced AI development.

\end{itemize}

\section{Related Works} \label{sec:relatedworks}
Complex reasoning tasks, such as math problems, have long been recognized as effective proxies for gauging the cognitive abilities of language models \cite{Sharples1989ComputersAT, KoncelKedziorski2016MAWPSAM, Szegedy2020APP, Polu2020GenerativeLM, Miao2020ADC, hendrycks2021measuring, cobbe2021training}. These tasks require the ability to understand symbols and text, dissect problems into logically connected sub-problems, combine results, and synthesize final solutions. They engage cognitive functions such as pattern induction, formula recall, deductive rule application, and abstract symbolic reasoning.

GSM8K \citep{cobbe2021training} and MATH \citep{hendrycks2021measuring} have been prominent benchmarks for evaluating the math reasoning capabilities of large language models (LLMs) in recent years. The chain of thought approach, proposed by \cite{Wei2022ChainOT}, addresses multi-step reasoning tasks by breaking them down into manageable steps.
Stanford Alpaca \citep{alpaca} has popularized the knowledge distillation method of cloning abilities from ChatGPT \citep{OpenAI2022CHATGPT} by generating related QA pairs. WizardMath \citep{luo2023wizardmath} further refined this distillation by specifying the difficulties of the QA pairs in the generation process. Mammoth \citep{Yue2023MAmmoTHBM} combines chain of thought and program of thought, finetuning its models with answers generated by GPT-4 \citep{OpenAI2023GPT4TR} in either natural or code language. MetaMath \citep{yu2023metamath} broadens the variety of question types by introducing forward/backward reasoning variations.

Despite significant advancements in math reasoning, evidence suggests that large language models may not fully master reasoning or even understand their own outputs. For example, \cite{Dziri2023FaithAF} found that LLMs fail to generalize to questions of varying complexity from their training data. \cite{DBLP:journals/corr/abs-2308-03762} demonstrated that, despite occasional analytical brilliance, GPT-4 is still severely limited in its reasoning capabilities. Similarly, \cite{Huang2023LargeLM} and \cite{Yen2023ThreeQC} have shown that ChatGPT struggles to judge the correctness of math problem solutions. However, our work focuses on constructing a qualitative and quantitative evaluation framework and discusses the evaluation principles and deficiencies of the current training paradigm in greater depth.

\section{Dataset Construction} \label{sec:dataset}

\begin{figure*}[ht]
\centering
\includegraphics[width=\textwidth]{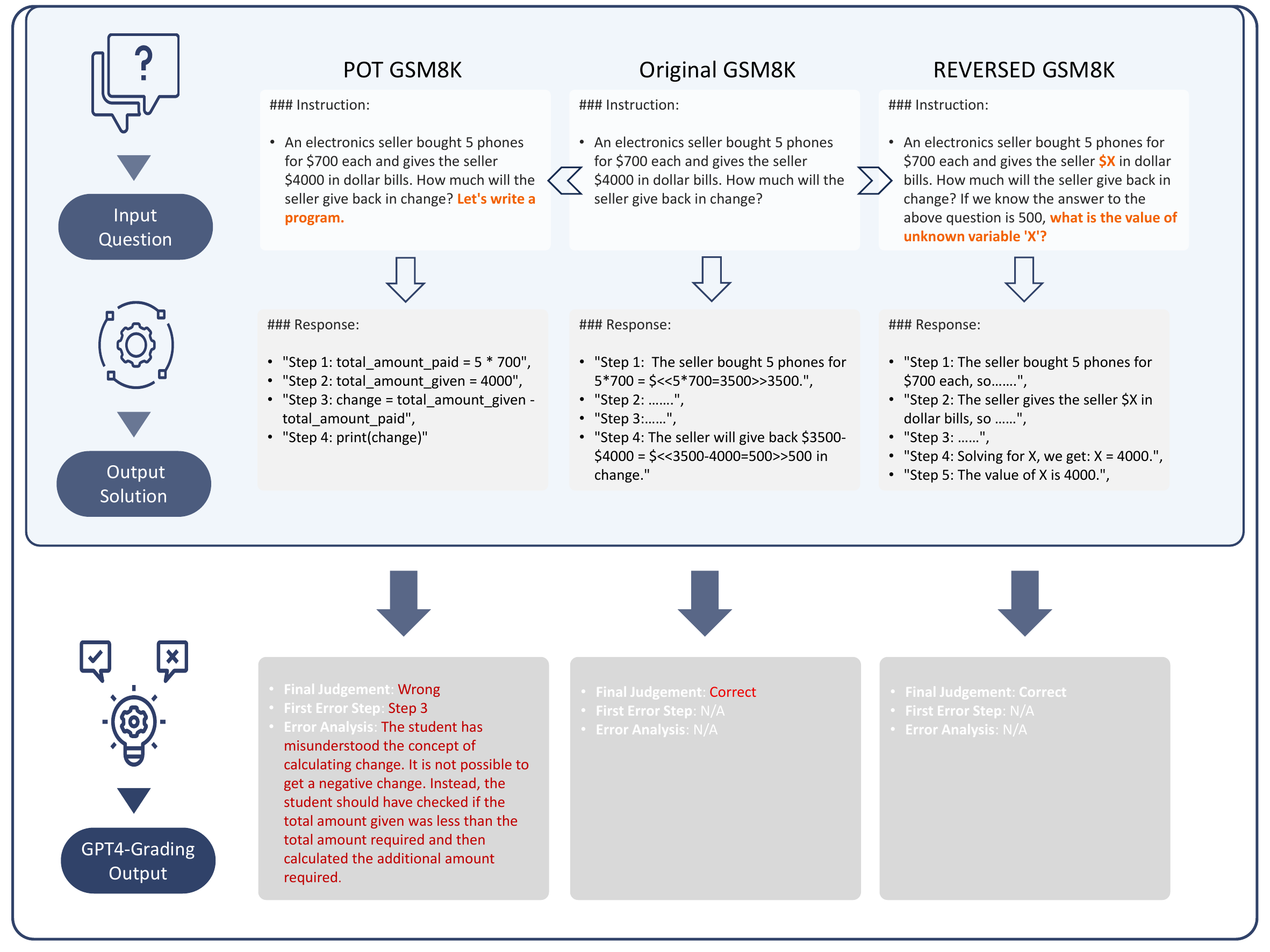}
\vspace{-0.6cm}
\caption{Structure of the MR-GSM8k benchmark and its evaluation paradigm.}
\label{fig:diagQuesType}
\end{figure*}

\subsection{Construction}
\textbf{Question Types:} The MR-GSM8k consists of three distinct types of questions. The first type includes original GSM8K instances, sampled directly from the dataset \cite{cobbe2021training}. The second type modifies GSM8K questions to include prompts requesting code solutions, as explored in \cite{Yue2023MAmmoTHBM}. The third type, termed reversed reasoning \cite{yu2023metamath}, involves concealing one of the inputs and asking for solutions that compute the missing input using the provided original answer. These variations were selected for their potential to broaden the scope of reasoning methodologies in LLMs. The "Program of Thought" approach, which includes code solutions, was proposed by \cite{Madaan2022LanguageMO} and has been empirically validated as an effective framework for math reasoning \cite{Yue2023MAmmoTHBM, Gou2023ToRAAT}. Reversed reasoning, a method that tests whether models can effectively learn backward relationships, has been recently emphasized in studies such as \cite{Berglund2023TheRC}, presenting additional challenges for these systems.

\textbf{Solution:} For each question collected, we used MetaMath-7B \cite{yu2023metamath} with a temperature setting of 1 to generate step-by-step solutions. We found this setting results in outputs with decent accuracy and nuanced mistakes that we would like the annotators and evaluated models to figure out. Intentionally, we targeted a result accuracy of approximately 50\% for the collected question-solution pairs, ensuring a balance between correct outcomes and those arising from flawed reasoning processes.

\subsection{Annotation}
For each collected question-solution pair, a panel of selected annotators was recruited to review each question-solution pair for its reasoning process and determine the following fields:

\textbf{Solution Correctness}: Solutions that yield a final output differing from the established ground truth are automatically marked as incorrect. However, in cases where the solution's final output aligns with the ground truth, annotators are tasked with reviewing the entire reasoning path. Their objective is to ascertain whether the correct output is the result of a logical and sensible reasoning process.

\textbf{First Error Step}: This attribute is applicable for solutions with either an unmatched final output or a matched final output underpinned by flawed reasoning. Annotators identify the initial step in the reasoning process where the logic deviates from correctness. In line with the approach of \cite{Lightman2023LetsVS}, we dissected GSM8K solutions into discrete steps, each marked by a newline character, and indexed them accordingly. Each step is then categorized as positive, neutral, or negative. Positive and neutral steps represent stages in the reasoning process where the correct final output remains attainable, whereas negative steps indicate a divergence from the path leading to the correct solution.

\textbf{Error Analysis}: Beyond identifying the first erroneous step, annotators are also responsible for conducting an in-depth analysis of the reasoning that led to the error. This involves an examination of the solution's reasoning flow, focusing on the cause of the initial error and what the correct line of reasoning should have been at that juncture. This error analysis is subsequently compared against the reasoning errors identified by the evaluated models during testing, to assess their accuracy and validity.

\subsection{Quality Control}
\textbf{Annotators:} Our panel of annotators is both well-trained and highly selective. Each candidate is required to thoroughly review our annotation manual (see Appendix-\ref{sec:appendix_anno_manual_examples} for details) and watch a demonstration video. Following this, candidates undergo a trial labeling process using a small, hold-out dataset. Recruitment decisions are based on performance during this trial. Additionally, four quality control supervisors, including the first author, are designated to address queries and monitor the labeling quality throughout the project.

\textbf{Annotation Procedure:} Every question in the MR-GSM8K undergoes multiple levels of scrutiny before being used in the evaluation process. Initially, each question is labeled twice by different annotators to ensure consistency. Any discrepancies in labeling, particularly regarding solution correctness or the identification of the first error step, are flagged and reviewed by a quality control supervisor. Furthermore, 50 percent of the labeled questions are randomly selected for a second round of quality control to identify and eliminate any inconsistent error steps or reasons. In the final stage of quality control, approximately 10 percent of the questions are manually inspected by the authors to ensure accuracy. This rigorous process ensures that any questions with incorrect error steps or reasoning are rectified before inclusion in the final dataset.

\begin{table}
\centering
\setlength\extrarowheight{2.5pt}
\caption{\data statistics: The first two columns are the numbers of correct and incorrect solutions. The  following two columns hold the lengths of questions and solutions, which are measured in number of words. The last two column shows the averaged solution steps and first error steps.}
\resizebox{\textwidth}{!}{
\begin{tabular}{lcccccc}
\hline
\textbf{Question Types} & \textbf{Correct} & \textbf{Incorrect} & \textbf{Q Length} & \textbf{S Length} & \textbf{S Steps} & \textbf{First Error Steps} \\ 
\hline
Original & 693 &  725 & 46.9 & 100.5 & 7. & 3.4 \\
POT & 113 & 109 & 45.1 & 34.9 & 6.5 & 3.9 \\
Reverse & 622 & 737 & 62.4 & 157.0 & 11.3 & 4.5 \\ 
Total & 1428 & 1571 & 51.5 & 97.5 & 8.3 & 3.9 \\ 
\hline
\end{tabular}
}
\label{tab:datasetStats}
\end{table}

\subsection{Dataset Statistics}
Table-\ref{tab:datasetStats} presents the statistics of \data, illustrating the distribution of correct and incorrect solutions across the three different types of questions. It is noteworthy that the reversed question type exhibits a significantly longer average question length due to its construction methodology. This type of question, due to its complex nature, also tends to have longer solution lengths as indicated in the table. Conversely, questions classified under the Program-of-Thought category, which typically require code solutions, have the shortest solution lengths, reflecting the concise and succinct nature of coding language. Despite these variations in average solution lengths and step counts, the average position of the first error step remains remarkably consistent across all question types.

\section{Evaluation Process \& Metric}
\textbf{Evaluation} As illustrated in Figure-\ref{fig:diagQuesType}, instead of simply solving a given question, the evaluated models are now presented with a question and a paired potential solution to score (e.g., the upper light blue part in the figure). Specifically, the evaluated model (e.g., the lower white part) is required to predict the correctness of the solution. If the solution is deemed incorrect, the model must further identify the first-error-step and explain the error-reason.
The solution-correctness and first-error-step are scored automatically based on comparison with manual annotations. Only when the evaluated model correctly identifies an incorrect solution and its first-error-step will its error-reason be further examined, either manually or automatically by models (see our discussions on error reason examination in the limitation section of Appendix-\ref{sec:limitations}). 

\textbf{Metrics} To provide a unified and normalized score reflecting the overall competence of the evaluated model, we propose a novel metric named \textbf{MR-Score}, consisting of three sub-metrics.
The first sub-metric is the Matthews Correlation Coefficient (MCC, \cite{Matthews1975ComparisonOT}) for binary classification of solution-correctness.
\begin{equation} \label{eq1}
\begin{aligned}
MCC = \frac{TP \times TN - FP \times FN}{\sqrt{(TP + FP) \times (TP + FN) \times (TN + FP) \times (TN + FN)}}
\end{aligned}
\end{equation}
Here, TP, TN, FP, and FN represent true positive, true negative, false positive, and false negative, respectively. The MCC score ranges from -1 to +1, where -1 indicates total disagreement between prediction and observation, 0 suggests performance near random, and +1 denotes perfect prediction. In the context of this paper, negative values are interpreted as no better than random guesses, and 0 is set as the cut-off threshold for normalization purposes.

The second metric is the accuracy of the first-error-step prediction:
\begin{equation} \label{eq2}
\begin{split}
ACC_{\text{step}} = \frac{N_{\text{correct\_first\_error\_step}}}{N_{\text{incorrect\_sols}}}
\end{split}
\end{equation}

The third metric calculates the accuracy of identifying both the correct first-error-step and the error-reason:
\begin{equation} \label{eq3}
\begin{split}
ACC_{\text{reason}} = \frac{N_{\text{correct\_error\_reason}}}{N_{\text{incorrect\_sols}}}
\end{split}
\end{equation}

\textbf{MR-Score} is a weighted combination of these three metrics:
\begin{equation} \label{eq4}
\begin{aligned}
MR\mbox{-}Score &= w_1 * \max(0, MCC) + w_2 * ACC_{\text{step}} 
 + w_3 * ACC_{\text{reason}} 
\end{aligned}
\end{equation}

The weights $w_1, w_2,$ and $w_3$ are chosen empirically to maximize differentiation between model performances by taking the difficulties of each task into account. For an extended discussion on the design of MR-Score, please refer to Appendix-\ref{sec:appendix_MRScore}.

\section{Experiments} \label{sec:experiments}


\begin{table*}[t]
\caption{Evaluation results on \data}
\setlength\extrarowheight{3.5pt}
\renewcommand{\arraystretch}{0.85}
\setlength{\tabcolsep}{3pt}  
\resizebox{\textwidth}{!}{
\begin{tabular}{lrrp{2mm}rrp{2mm}rrp{2mm}rrp{2mm}rrp{2mm}rr}  
\toprule
\multirow{2}{*}{\textbf{Model}} & \multicolumn{2}{c}{\textbf{Task1-TPR}} && \multicolumn{2}{c}{\textbf{Task1-TNR}} &&  \multicolumn{2}{c}{\textbf{Task1-MCC}} && \multicolumn{2}{c}{\textbf{Task2-Accy}} && \multicolumn{2}{c}{\textbf{Task3-Accy}} && \multicolumn{2}{c}{\textbf{MR-Score}} \\ 
\cline{2-3} \cline{5-6} \cline{8-9} \cline{11-12} \cline{14-15} \cline{17-18}
& $k$=0 & $k$=3 && $k$=0 & $k$=3 && $k$=0 & $k$=3 && $k$=0 & $k$=3 && $k$=0 & $k$=3 && $k$=0 & $k$=3 \\ 
\midrule

\multicolumn{18}{c}{\textbf{Open-Source Small}}\\
\midrule
Qwen-1.8B & 21.8 & 33.3 && 0.1 & 3.9 && 0. & 0. && 0. & 0.4 && 0. & 0. && 0. & 0.1 \\
Phi3-3.8B & 11.3 & 62.6 && 98.5 & 72.6 && 20.4 & 35.4 && 32.9 & 26.3 && 18.0 & 13.9 && \textbf{22.9} & 21.9 \\
\midrule
\multicolumn{18}{c}{\textbf{Open-Source Medium}}\\
\midrule
Deepseek-Math-7B-RL & 77.3 & 2.4 && 52.3 & 0.4 && 30.4 & 0. && 9.8 & 0.1 && 5.1 & 0.1 && 11.6 & 0.1 \\
WizardMath-v1.1-7B & 99.3 & 6.7 && 0.5 & 0.6 && 0.0 & 0.0 && 0.3 & 0.2 && 0.3 & 0.1 && 0.2 & 0.1 \\
Llama3-8B & 3.2 & 40.9 && 98.3 & 80.3 && 5.1 & 23.1 && 29.1 & 23.3 && 15.0 & 11.6 && 17.2 & \textbf{17.4} \\
\midrule
\multicolumn{18}{c}{\textbf{Open-Source Large}}\\
\midrule
MAmmoTH-70B & 88.0 & 89.8 && 23.1 & 2.8 && 14.6 & 0.0 && 3.9 & 0.3 && 1.8 & 0.3 && 5.0 & 0.2 \\
MetaMath-70B & 7.8 & 0.0 && 0.3 & 0.0 && 0.0 & 0.0 && 0.1 & 0.0 && 0.0 & 0.0 && 0.0 & 0.0 \\

Llama3-70B & 67.6 & 89.3 && 83.0 & 66.0 && 51.3 & 56.4 && 38.9 & 33.5 && 32.7 & 25.7 && \textbf{38.3} & 34.2 \\
Qwen1.5-72B & 83.7 & 87.7 && 57.1 & 52.4 && 42.0 & 42.5 && 19.1 & 23.1 && 13.5 & 15.8 && 20.9 & 23.3 \\
Deepseek-v2-236B & 60.1 & 88.2 && 87.2 & 61.5 && 49.4 & 51.2 && 26.8 & 32.4 && 23.8 & 28.3 && 29.8 & 34.1 \\
\midrule
\multicolumn{18}{c}{\textbf{Closed-Source LLMs}}\\
\midrule
Claude3-Haiku & 70.4 & 99.0 && 51.7 & 8.1 && 22.5 & 16.7 && 17.2 & 2.3 && 11.3 & 1.8 && 15.3 & 4.9 \\
GPT-3.5-Turbo & 16.3 & 59.7 && 93.8 & 65.7 && 16.2 & 25.5 && 30.6 & 21.0 && 20.3 & 13.0 && 22.6 & 17.9 \\
Claude3-Sonnet & 35.1 & 88.4 && 89.8 & 44.8 && 30.0 & 36.5 && 25.2 & 18.8 && 19.9 & 15.6 && 23.5 & 20.8 \\

GPT-4-Turbo & 69.5 & 83.0 && 91.8 & 84.2 && 63.3 & 67.2 && 48.8 & 51.7 && 46.3 & 48.1 && 50.5 & \textbf{53.0} \\

\bottomrule
\end{tabular}
}
\label{tab:main_table}
\vspace{-2mm}
\end{table*}

\subsection{Experiment Setup}
To evaluate the performance of different language models on our benchmark, we selected models from diverse backgrounds. These models vary greatly in size, ranging from a few billion parameters, such as Qwen-v1.5-1.8B \cite{DBLP:journals/corr/abs-2309-16609}, to 70 billion parameters like Llama3-70B \cite{touvron2023llama}, and up to 236 billion parameters as seen in Deepseek-v2-236B \cite{DBLP:journals/corr/abs-2401-02954}. Additionally, to contrast performances between models fine-tuned from general instructions and those specialized in math problems, we included representative math models from the open-source community, such as WizardMath-v1.1-7B \cite{luo2023wizardmath}, MAmmoTH-70B \cite{Yue2023MAmmoTHBM}, DeepseekMath-7B-RL \cite{DBLP:journals/corr/abs-2402-03300}, and MetaMath-70B \cite{yu2023metamath}. Furthermore, to explore the differences between commercial and open-source models, we included models from the OpenAI GPT family \cite{OpenAI2022CHATGPT} and the Anthropic Claude-3 series \cite{ANTHROPIC2023CLAUDE}.

Each model was evaluated under a zero-shot setting to assess their ability to follow instructions and their mathematical reasoning capabilities. Given that some evaluated models are not fine-tuned for general instruction following, we also tested each model under a few-shot setting to leverage their in-context learning abilities for understanding mathematical reasoning (see our prompts in Figure-\ref{fig:zero_shot_prompt} and \ref{fig:few_shot_prompt} in the Appendix). To ensure reproducibility and minimize variance, the inference temperature was set to zero across all models.
\subsection{Experiment Results}
Our evaluation results are presented in Table-\ref{tab:main_table}, where tasks 1, 2, and 3 correspond to determining solution correctness, the first error step, and the error reason, respectively. For Task 1, we also report the true positive rate and true negative rate. Key observations from our study are as follows:

\textbf{Overall Performance:} As depicted in the table, GPT-4-Turbo significantly outperforms all other models across both open-source and closed-source domains. Among the open-source models, Llama3-70B exhibits the closest performance to GPT-4, yet it still lags by more than 12 absolute points in MR-Score. In the small to medium model size category, Phi3-3.8B outshines others, even surpassing the scores of Claude3-Haiku. Notably, most specialized models we evaluated failed to adapt to our evaluation paradigm and systematically underperformed compared to the generalized chat models. Despite similar levels of success on GSM8K, as illustrated in Figure-\ref{fig:gsm_vs_mrgsm}, all models tested show a significant drop in performance in our benchmark, resulting in a much wider differentiation in scores.

\textbf{Performance by Model Size:}
Contrary to the common belief that larger models inherently possess greater capabilities, our findings challenge this notion. Specifically, Phi3-3.8B performed substantially better than other models at the 7B level and achieved comparable performance with Qwen1.5-72B, which is approximately twenty times larger. A similar trend is observed between Llama3-70B and Deepseek-v2-236B, with Llama3-70B outperforming the latter in both zero-shot and few-shot settings. These results suggest that while model size is an important factor in reasoning ability, the quality of pretraining data and the application of data synthesis techniques such as knowledge distillation may also play crucial roles.

\textbf{Specialized Math Models:}
Within the open-source community, a multitude of models are dedicated to math reasoning, employing various fine-tuning techniques and datasets. Despite this, most models failed significantly in our benchmark. Specifically, WizardMath-v1.1-7B and MetaMath-70B appeared overfitted to the GSM8K response format and were unable to adhere to our evaluation instructions, both with and without few-shot demonstrations. Conversely, Deepseek-Math-7B-RL and MAmmoTH-70B managed to comprehend our complex evaluation instructions and achieved decent performance.

\textbf{Few Shot Demonstrations:}
Given that the MR-GSM8K benchmark poses significant challenges in complex instruction-following in addition to mathematical reasoning, we explored whether providing few-shot demonstrations could enhance the performance of specialized models. However, as indicated in Table-\ref{tab:main_table}, the few-shot setting proved detrimental to all tested models. Although Deepseek-Math-7B-RL and MAmmoTH-70B demonstrated decent performance in the zero-shot setting, they struggled to adhere to the desired task instructions in the few-shot setting, reverting to the question-answering paradigm when presented with extended context and demonstrations. Similarly, few-shot examples did not aid WizardMath-v1.1-7B and MetaMath-70B in adhering to the expected scoring and reasoning format. These specialized models exhibited a strong tendency to revert to the question-answering paradigm, an issue not as prevalent in general chat models. This tendency suggests that fine-tuning on a narrowly focused dataset, often sampled or augmented from specific math datasets like GSM8K, may lead to overfitting to a particular input/output data distribution, resulting in only a superficial mastery of mathematical reasoning. Despite the close similarity between MR-GSM8K and the datasets these models were exposed to, their underperformance highlights a critical shortcoming in the generalization capabilities of reasoning abilities developed through specialized fine-tuning.

For general chat models, the impact of few-shot demonstrations varied significantly across models, even within the same family, with outcomes ranging from slight improvements to notable deteriorations. No consistent pattern emerged, indicating that improvements do not uniformly correlate with model size or initial benchmark performance.

\textbf{In Context Learning Bias:} An intriguing outcome from our few-shot experiments is the significant impact this setting had on the models' propensity to score solutions as correct or incorrect. Analyzing the true positive rate (TPR) and true negative rate (TNR) in the solution correctness task, we empirically observed that most models exhibited an increase in TPR while concurrently showing a reduction in TNR under the three-shot setting. This trend can likely be attributed to the composition of our few-shot examples, which included two correct solutions and one incorrect solution, suggesting that the distribution of correctness within these examples may influence model predictions. To verify our conjecture, we conducted an ablation study on the number of correct solutions and the result shows strong correlation with our proposal (see Figure-\ref{fig:kshot-tpr-tnr} in Appendix for more details). This susceptibility of language models to the distribution of few-shot examples highlights a fundamental flaw in the current reasoning paradigms: rather than making scoring decisions based purely on reasoning, the models appear to be swayed by the few-shot examples. This influence might also account for the generally worse performance observed in the few-shot experiments compared to the zero-shot settings, where such biases are absent.

\begin{figure}[ht]
    \centering
    \includegraphics[width=\textwidth, trim=0cm 7cm 0cm 6.2cm, clip=true]{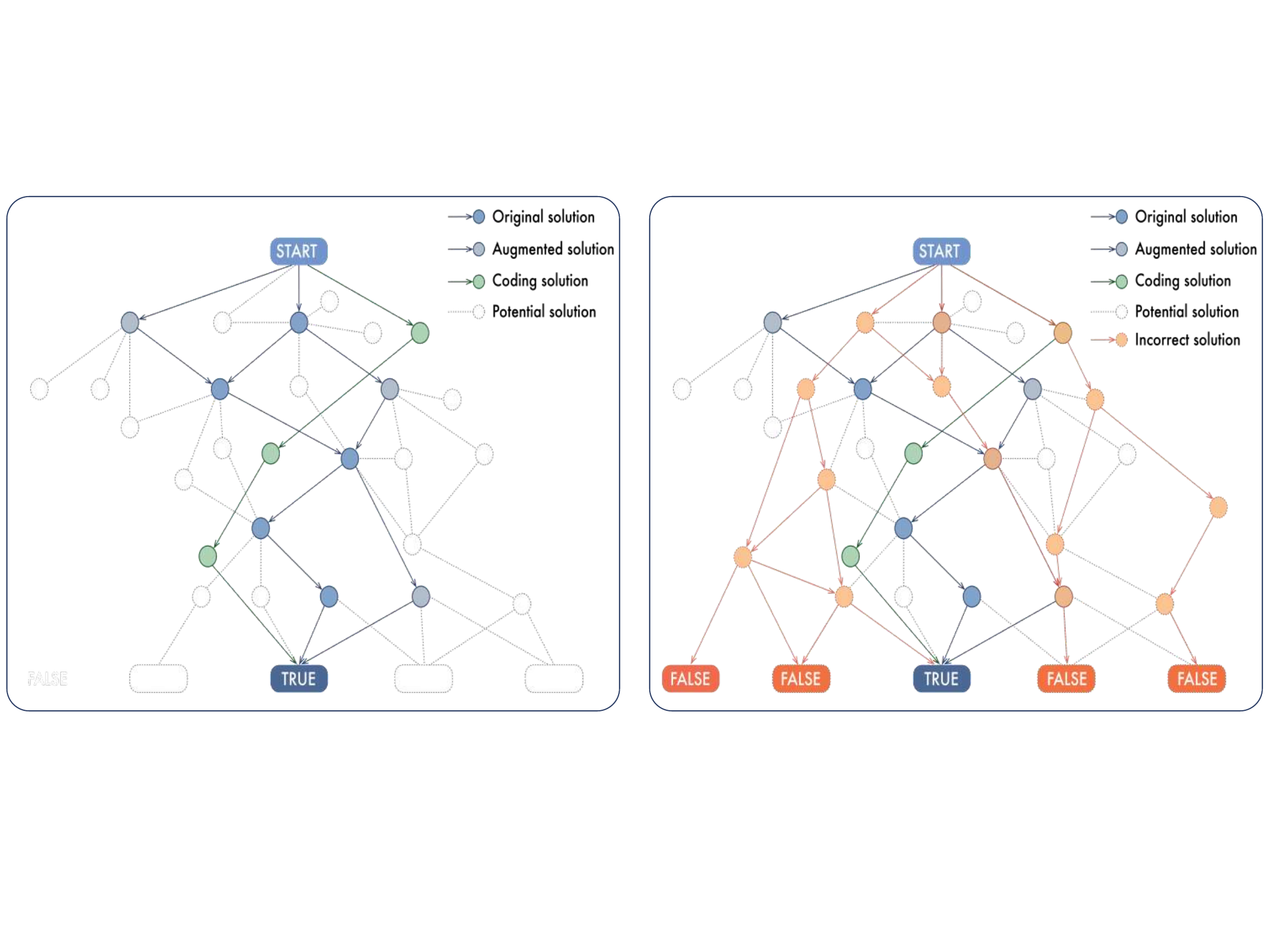}
    \vspace{-0.7cm}
    \caption{This figure aims to illustrate the fundamental limitations of the current training and evaluation paradigms for LLMs. The left side of the figure demonstrates that utilizing only correct solutions limits coverage of the solution space despite various data augmentation techniques. The right side depicts potential solutions that might contain incorrect steps or reach the final computation result through an incorrect reasoning path. Models trained exclusively on correct reasoning paths often fail to assess the validity or discern nuanced differences of alternative reasoning paths for the same problem. This highlights a critical gap in the training paradigm, where models exhibit basic imitation skills but lack a deep understanding of the underlying logical rules, leading to a superficial grasp of reasoning processes.}
    \label{fig:ReasoningPath}
\end{figure}

\section{Discussion} \label{sec:discussion}
\subsection{Case Studies on Reasoning Failures}
In Appendix-\ref{sec:appendix_anno_manual_examples}, we provide several case studies of responses generated by evaluated models. These examples help illuminate various reasoning errors that are often obscured in simpler benchmarks such as GSM8K. First, models frequently exhibit what is known as the "reversal curse," where they mistakenly claim that "A equals to B" is not equivalent to "B equals to A" \cite{Berglund2023TheRC}. Second, inconsistencies and errors in quantity unit usage by the models suggest a fundamental lack of ontological understanding regarding the properties of the quantities they manipulate. Third, many models display insensitivity to numerical computation errors, highlighting an inherent weakness in arithmetic processing by language models. These observations underscore the critical gaps in model training and evaluation, pointing to an over-reliance on correct solution paths that neglects the broader context and complexity of real-world reasoning.

\subsection{What Is the Significance of Reason About Reasoning?} 
In this paper, we have demonstrated that simply observing computation results is insufficient to uncover the cognitive depth of evaluated models. Equally important is the validity and logic of the reasoning process employed by these models. For a model to successfully diagnose solution correctness, it must infer the correct result and also engage in counterfactual reasoning along different reasoning paths, actively examining the conditions and assumptions made at various steps. Success in this paradigm is unlikely without a holistic understanding and robust mastery of the underlying concepts. Thus, the "reason about reasoning" paradigm emerges as a critical meta-evaluative tool.

Another key significance of this paradigm is its capability to transform any existing benchmark into a more robust and holistic assessment tool. As highlighted by \cite{Balloccu2024LeakCR} and \cite{Yang2023RethinkingBA}, data contamination issues are becoming increasingly prevalent and elusive to detect. Our paradigm not only facilitates modifications to existing benchmarks but also demonstrates robustness against potential data contamination, as evidenced by our experiments across a wide array of state-of-the-art LLMs.

\section{Conclusion}
Throughout this paper, we have explored the inadequacies of prevalent math reasoning benchmarks and introduced a pioneering evaluation paradigm that compels models to engage in meta-reasoning. Our empirical findings demonstrate that this novel paradigm enables our benchmark to effectively differentiate between models and uncover their various deficiencies. This differentiation has been particularly evident in the performance struggles of state-of-the-art language models when confronted with our benchmark, revealing significant shortcomings in current training methodologies.

These revelations underscore the need for a critical reevaluation of existing training and evaluation practices in the realm of large language models. By advocating for the widespread adoption of our "reason about reasoning" evaluation paradigm, we encourage researchers to adapt and broaden other reasoning benchmarks similarly. Such transformation is vital not only for a more rigorous assessment of LLMs but also for fostering a deeper and more holistic understanding of these models' capabilities.
\clearpage 
\bibliographystyle{plainnat}
\bibliography{custom}
\clearpage

\appendix
\begin{table}
\centering
\begin{tikzpicture}
\matrix (confusion) [matrix of nodes, nodes in empty cells,
  column sep=10pt, row sep=10pt,
  nodes={minimum height=15mm, minimum width=15mm, draw=black, line width=1pt, anchor=center},
  row 1/.style={nodes={draw=none, fill=none, minimum height=5mm}},
  column 1/.style={nodes={draw=none, fill=none, minimum width=5mm}}] {
  & Pos & Neg \\
Pred-Pos & 960/1042 & 218/626 \\
Pred-Neg & 82/1042 & 408/626 \\
};
\end{tikzpicture}
\caption{This confusion matrix represents the accuracy of GPT4-Turbo-1106 in assessing 1668 incorrect solutions that were correctly identified with the right error step. The task for GPT4-Turbo-1106 was to evaluate the correctness of the error reason provided by the evaluated model, in comparison with the actual ground truth labelled by expert. 'Pos' and 'Neg' represent the ground truth correctness of the provided explanation, while 'Pred-Pos' and 'Pred-Neg' indicate GPT4's prediction about the correctness.}
\label{tab:confusionMatrix}
\end{table}

\begin{figure*}[ht]
\centering
\includegraphics[width=\textwidth]{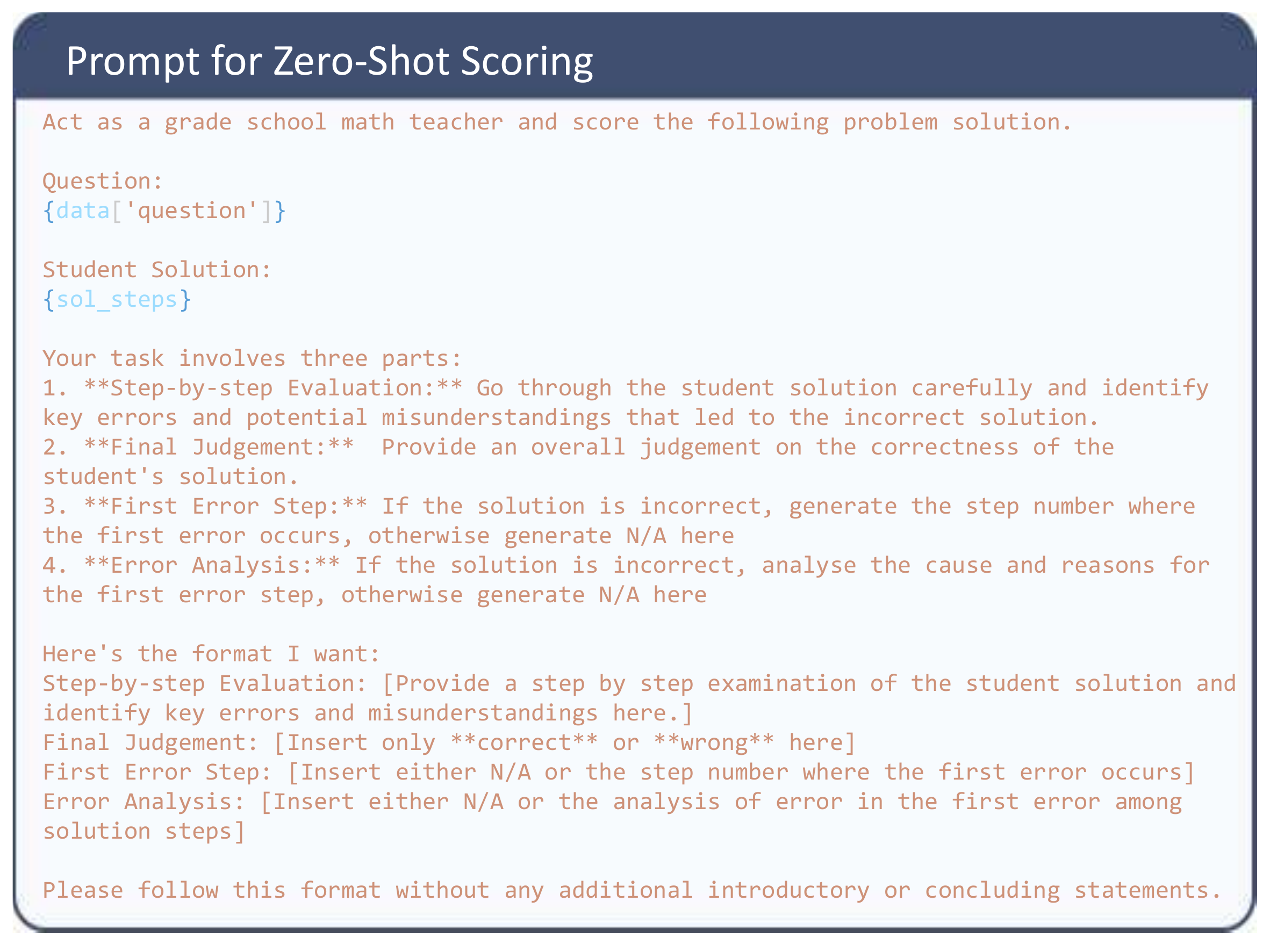}
\caption{This is the zero shot prompt we used to evaluate all the models}
\label{fig:zero_shot_prompt}
\end{figure*}

\begin{figure*}[ht]
\centering
\includegraphics[width=\textwidth]{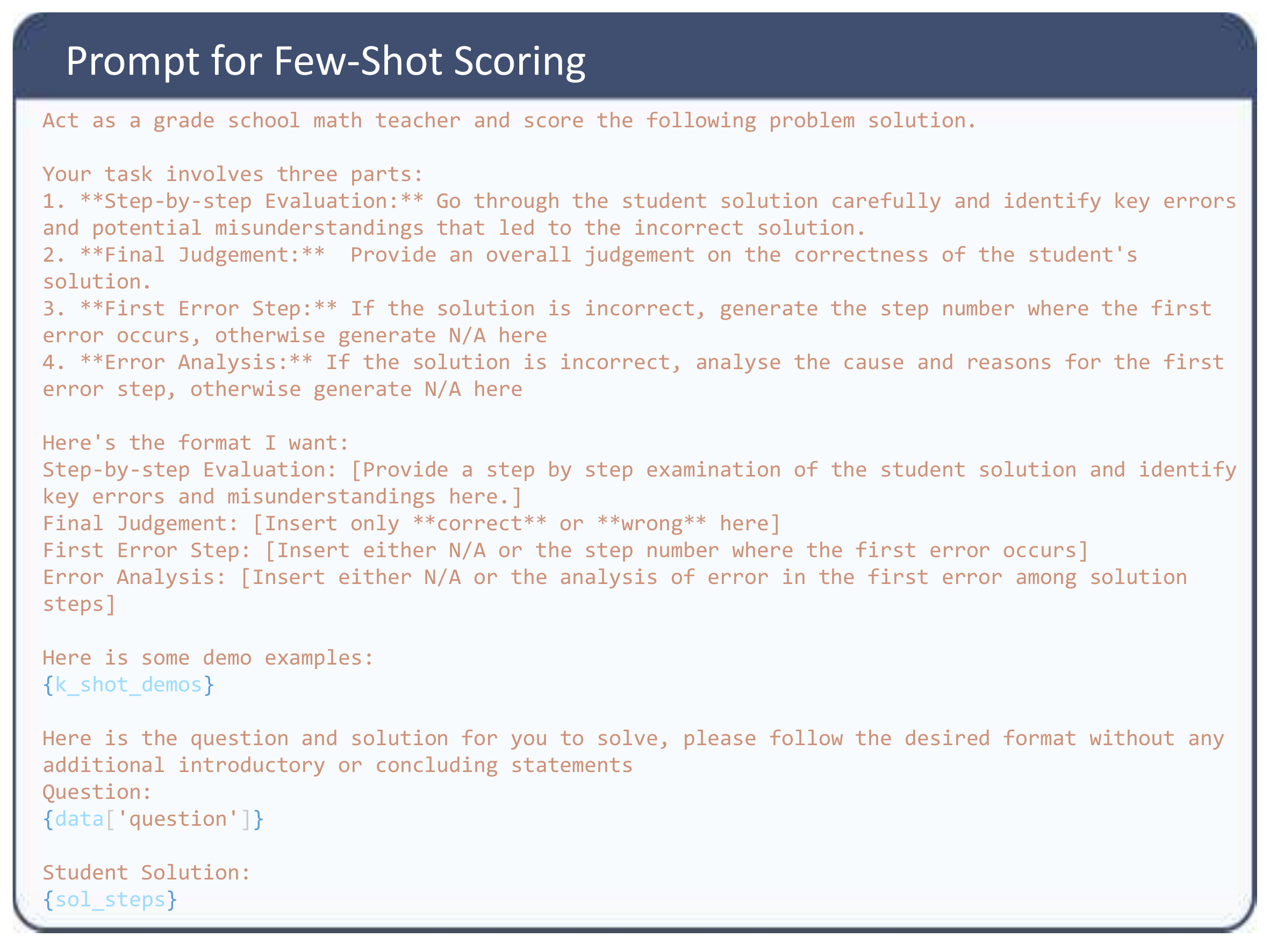}
\caption{This is the few shot prompt we used to evaluate all the models}
\label{fig:few_shot_prompt}
\end{figure*}

\begin{figure*}[ht]
\centering
\includegraphics[width=\textwidth]{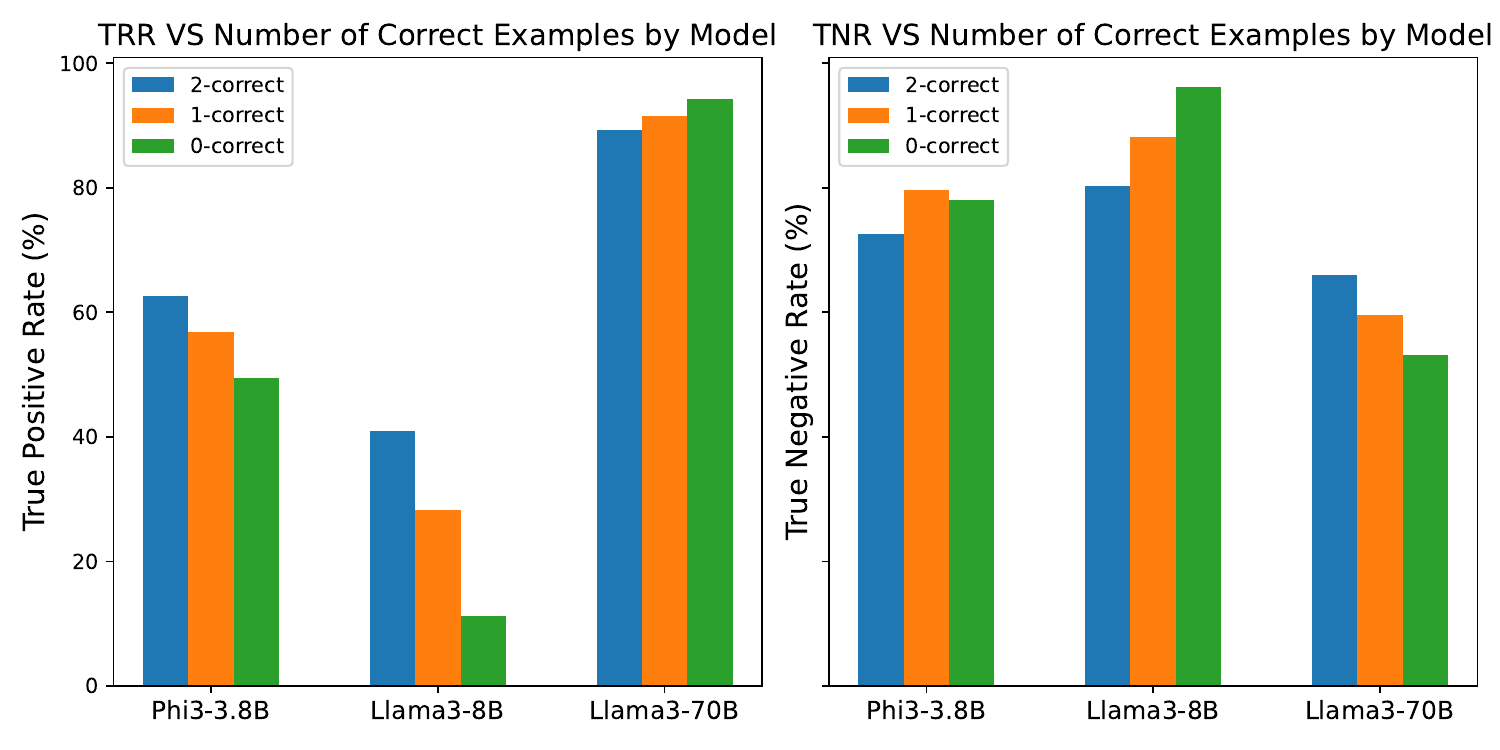}
\caption{This is how the true positive and true negative rates change with the number of correct solutions in the 3-shot demonstrations. Note for the smaller models like Phi-3 and Llama3-8B, their susceptibility trend is opposite with that of Llama3-70B.}
\label{fig:kshot-tpr-tnr}
\end{figure*}

\section{Limitations} \label{sec:limitations}
\textbf{Does MR-GSM8k Require Human Labeling?}

In this study, we proposed the MR-Score as a unified and normalized metric to evaluate the language models. The MR-Score consists of three sub-metrics, with the error reason being just one of the evaluation criteria. Similar to translation tasks, where expressions in one language may correspond to many variations in another, it is challenging to develop an automatic evaluator that scores the error reason perfectly. Despite these limitations, this does not undermine the arguments we present, nor does it affect the cognitive deficiencies unveiled by this metric. To the best of our knowledge, GPT-4 has been the most popular choice for serving as an automatic evaluator across different metrics \citep{Zheng2023JudgingLW, Liu2023AlignBenchBC}. In Appendix-\ref{sec:appendix_MRScore}, we empirically demonstrate that GPT-4 can serve as a decent automatic evaluator, with the final MR-Score based on its labeling results closely matching those of manual labeling.

\textbf{Limitations of the Meta Reasoning Evaluation Paradigm and MR-GSM8k Dataset}

Reflecting on Goodhart's law, which states that "When a measure becomes a target, it ceases to be a good measure," it's evident that the "reason about reasoning" paradigm is not immune to this phenomenon. This paradigm, like any other, can be targeted for optimization. However, our evaluation paradigm presents a greater challenge to overfitting compared to others, due to its demand for a comprehensive understanding within a broad error space, as illustrated in Section-\ref{sec:indomainfinetuning}.

On the other hand, the meta-reasoning evaluation framework in \data, while innovative, is not without its limitations. Firstly, its applicability may be restricted when it comes to subjects that are inherently holistic or creative in nature, such as humanities or sociology. These subjects often require a comprehensive understanding and modification (e.g. essay writing), which can be challenging to break down into specific, sequential reasoning steps and corrections. Secondly, \data is currently confined to questions in English. This could potentially limit the scope of reasoning challenges that can be explored, as different languages may present unique cognitive and linguistic hurdles. Lastly, the analysis and correction of errors in the reasoning steps are currently based on solutions generated by MetaMath-7B model only. It's important to note that different LLMs and different individuals, may exhibit distinct reasoning and error patterns. Therefore, it would be beneficial to broaden the spectrum of solutions analyzed, incorporating a more diverse range of LLMs and even human responses. This would not only enhance the robustness of the evaluation framework but also provide a more nuanced understanding of the reasoning processes at play.

\begin{figure*}[ht]
    \centering
    \includegraphics[width=\textwidth]{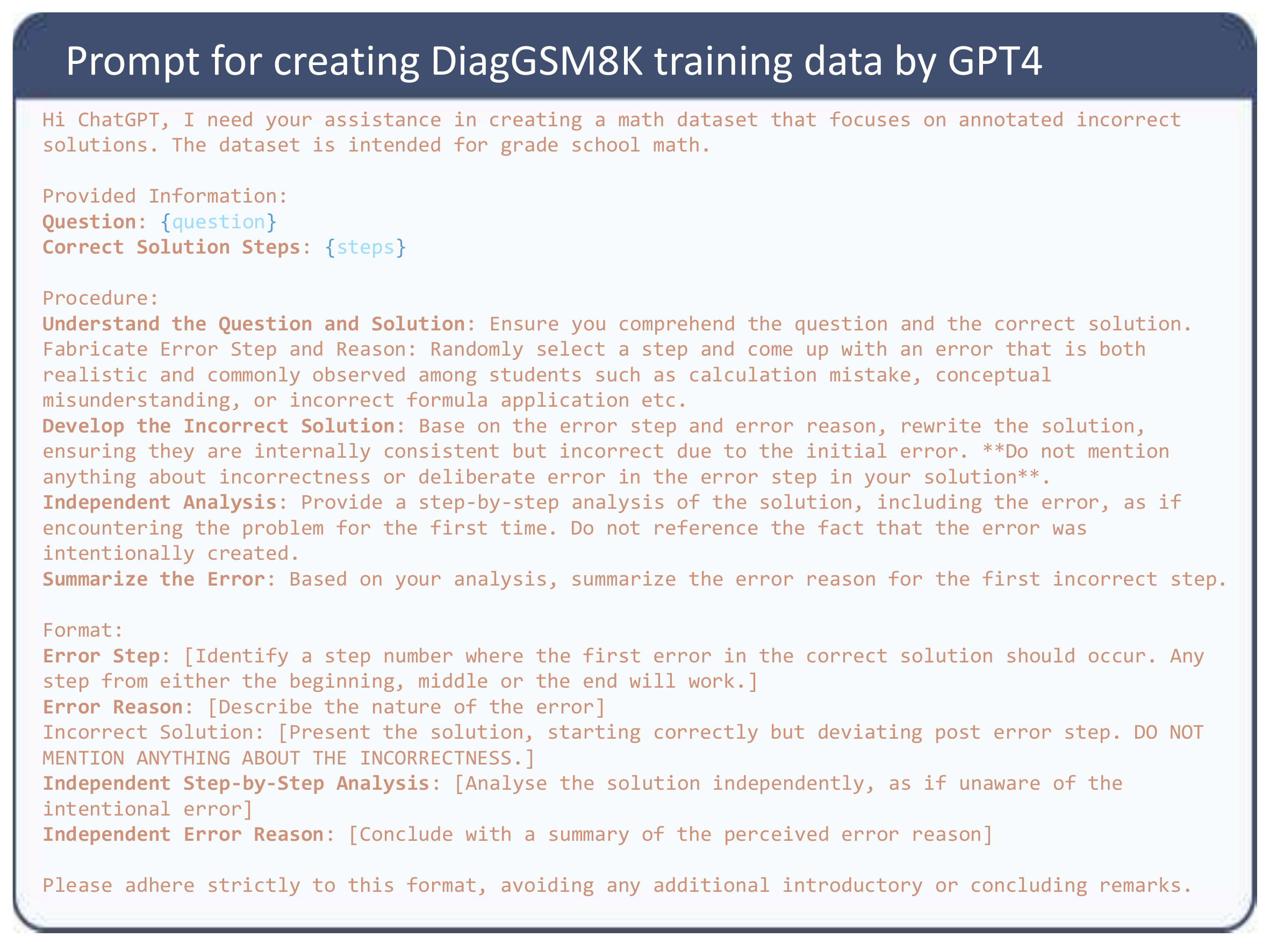}
    \caption{This is the prompt we used for GPT4 to create in-domain training data.} 
    \label{fig:TrainingGeneration}
\end{figure*}

\begin{table*}
\centering
\begin{tabular}{lccccc}
\hline
\textbf{Models} & \textbf{Step} & \textbf{Step+Reason/M} & \textbf{Step+Reason/A} & \textbf{MR-Score/M}  & \textbf{MR-Score/A} \\ 
\hline
GPT4 & 823/1573 & 677/1573 & 732/1573 & 0.495 & 0.512 \\
Claude2 & 331/1573 & 185/1573 & 224/1573  & 0.191 & 0.203 \\
llama2-70B-MR & 327/1573 & 99/1573 & 139/1573 & 0.105 & 0.118 \\
GPT3.5 & 179/1573 & 73/1573 & 73/1573 & 0.097 & 0.097 \\
MetaMath-70B & 22/1573 & 6/1573 & 7/1573 & 0.013 & 0.013 \\
Mammoth-70B & 4/1573 & 1/1573 & 2/1573 & 0.012 & 0.012 \\
WizardMath-70B & 6/1573 & 1/1573 & 1/1573 & 0.001 & 0.001 \\
\hline
\end{tabular}
\caption{Comparison of the manual labelling results and GPT4-Turbo-1106 labelling results. Step column shows the number that each evaluated models successfully located the first error steps among incorrect solutions. Step+Reason/M stands for the manual labelling results of the error reasons where its first error step is correct. 
Step+Reason/A corresponds to the labelling results of GPT4-Turbo-1106. llama2-70B-MR are llama2-70B model finetuned on the GSM8k training set and its meta-reasoning augmentation by GPT4. 
}
\label{tab:GPT4Score}
\end{table*}

\section{In Domain Finetuning} \label{sec:indomainfinetuning}

Given the challenges posed by the novel "reason about reasoning" task paradigm, we explored how much targeted task-specific training data could enhance the performance of current state-of-the-art models on this task. We considered augmenting the GSM8K training set with diagnostics data in a similar format. However, due to the labor-intensive nature of manual annotation, we opted for a more feasible approach using an expert-designed procedure where GPT-4 generates the training data based solely on the original GSM8K problems, excluding any Program of Thought (POT) or reversed transformations.

This process involved presenting GPT-4 with a question and its correct solution, then instructing it to introduce an error at a randomly chosen step and complete the solution accordingly. The step-by-step analysis was subsequently generated, focusing on the fabricated error. 
Despite GPT-4's modest 40+\% accuracy in correctly identifying incorrect solutions in the test set, this procedure successfully generated accurate diagnostic training data with over 90 percent accuracy, as verified by expert annotation on a sub-sampled set. This is largely due to the expert designed procedure (Figure-\ref{fig:TrainingGeneration}) that greatly lowered the difficulties for instruction following. Note, occasionally GPT4 will fail to fabricate a valid error due to the lack of true understanding of errors (e.g., switch the fraction 8/3 to 2 and 2/3 then claims this is an error).

For our base model, we used llama-2-70B-base, consistent with the approach of other open-source SOTA math models. We merged the GSM8K training set with the GPT-4 generated diagnostic data, consisting of 5k incorrect solutions and 4k correct solutions. For fine-tuning, we employed the Qlora method \citep{Dettmers2023QLoRAEF}, maintaining the same hyperparameters as used for MetaMath-70B. The evaluated results indicate a 31.74\% true positive rate and 73.49\% true negative rate, which lead to a 5.8\% score in MCC. The accuracy for first-error-step and error-reason is 20.79\% and 6.29\% respectively. The MR-Score for this finetuned model is 10.5\%.

\begin{figure}[t]
    \centering
    \includegraphics[width=\columnwidth]{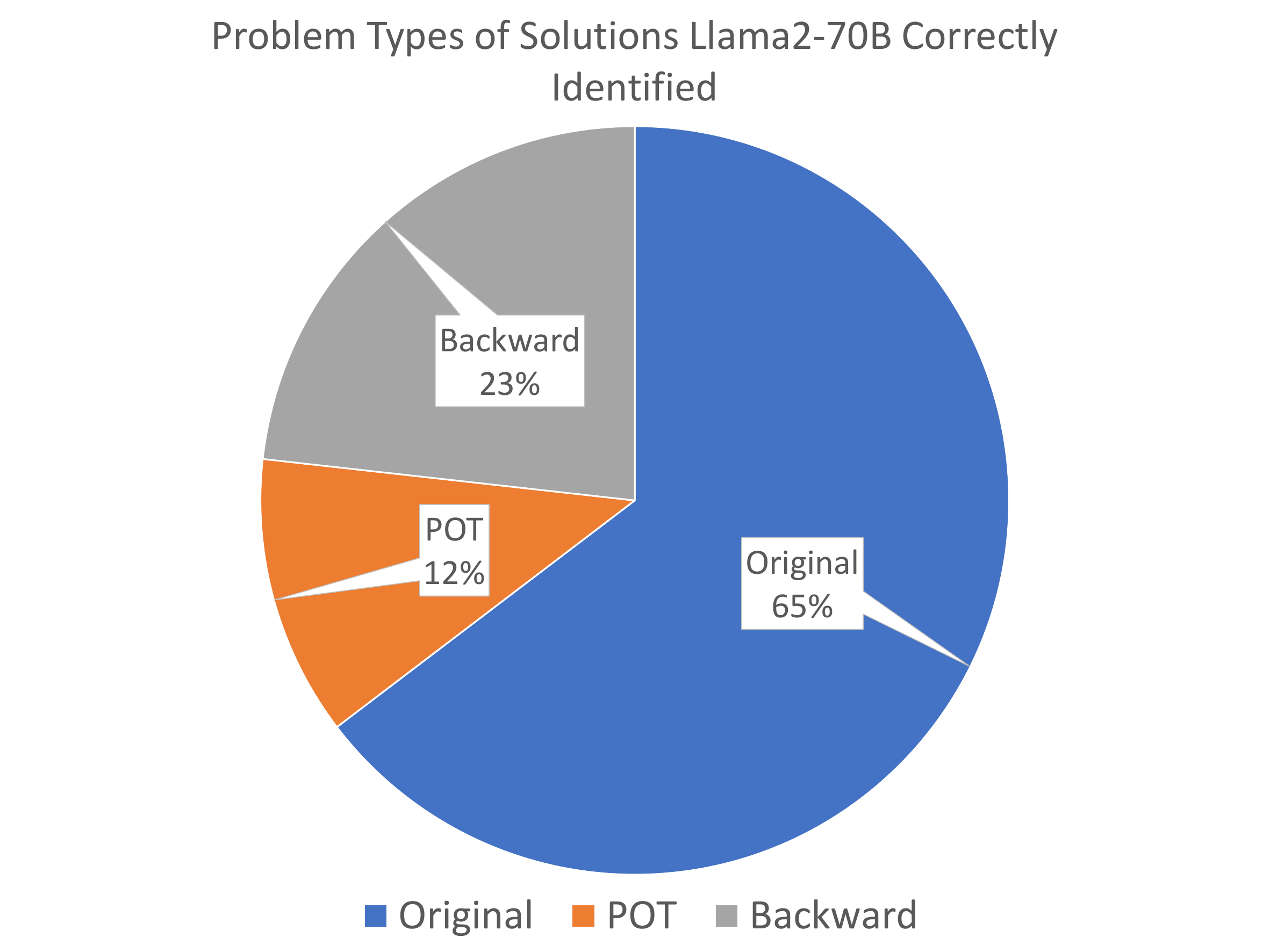}
    \vspace{-0.8cm}
    \caption{Problem types of incorrect solutions for which the llama2-70B-MR model has successfully identified both the error step and error reason. Note that the training set only included solutions from original problems.}
    \label{fig:problem_types}
\end{figure}

Notably, the fine-tuned Llama2 model demonstrated a distinct tendency from that of GPT3.5 and other open-source models; it was less inclined to accept solutions uncritically, tending instead to over-reject solutions regardless of their correctness. As depicted in Figure-\ref{fig:problem_types}, of the 99 questions where the model accurately predicted both correctness and the first error step, a significant portion involved questions with POT and reversed reasoning types. This is particularly noteworthy given that the model was trained exclusively on original questions.

Caution is necessary when interpreting the outcomes of in-domain fine-tuning. Although the fine-tuned model achieved improved results, it is important to recognize that the overall number of correct diagnoses for incorrect solutions remains relatively low (e.g., 6.29\%). This underscores the challenging nature of our MR-GSM8k benchmark, where effective diagnosis across diverse solution spaces requires a comprehensive understanding of the problem. Consequently, simple fine-tuning strategies may not yield substantial improvements in performance.

\section{Design thinking of MR-Score}
\label{sec:appendix_MRScore}
The MR-Score is consist of three sub-metrics corresponds to the three sequential reasoning sub-tasks. For the first solution correctness prediction, we empirically noticed that most of the evaluated language models tend to either blindly classify the given solution as correct or incorrect, exemplified by the low true-positive/true-negative rate in Table-\ref{tab:main_table}. Therefore, we chose the MCC score instead of metrics like F1 or Balanced-Accuracy due to its value range. The models that have high true-positive rate but low true negative rate will have near zero score under the MCC metric and vice versa. For the second and third tasks of locating first error step and elucidating error reason, we chose the simple accuracy metric. One of the reason is that locating the first error step is a multi-class classification problem and it is difficult to have large prediction bias while at the same time scores high accuracy. Similarly, the explaining error reason task is a free-form generation task that requires substantial understanding and a simple accuracy metric is enough to categorize the model behavior.       

As to the weights given to the three metrics, they are crafted by considering the task complexity and the difference between manual labelling and auto labelling results of the error reason. As discussed in Section-\ref{sec:limitations}, we chose the GPT-4-Turbo as our proxy evaluator and Table-\ref{tab:GPT4Score} is the results of auto-labelling vs our expert manual-labelling. It is clear that the final MR-Score calculated from manual labelling VS auto labelling are very close to each other, exhibiting the potential of GPT4 to serve as a delegate evaluator for our task.

Table-\ref{tab:confusionMatrix} displays the confusion matrix based on GPT4's labelling of all the error reasons. Notably, GPT4 is able to achieve 82\% of overall accuracy despite a substantial higher false positive rate than false negative rate. However, we still encourage large tech companies, who have the resources to bear manual labelling costs, to release open-source manual labelling results when publishing findings using MR-GSM8k for the best of rigorousness.

\section{Annotation Manuals, Examples and Case Studies}
\label{sec:appendix_anno_manual_examples}

\phantomsection
\label{pdf:anno_materials}
\includepdf[pages=-]{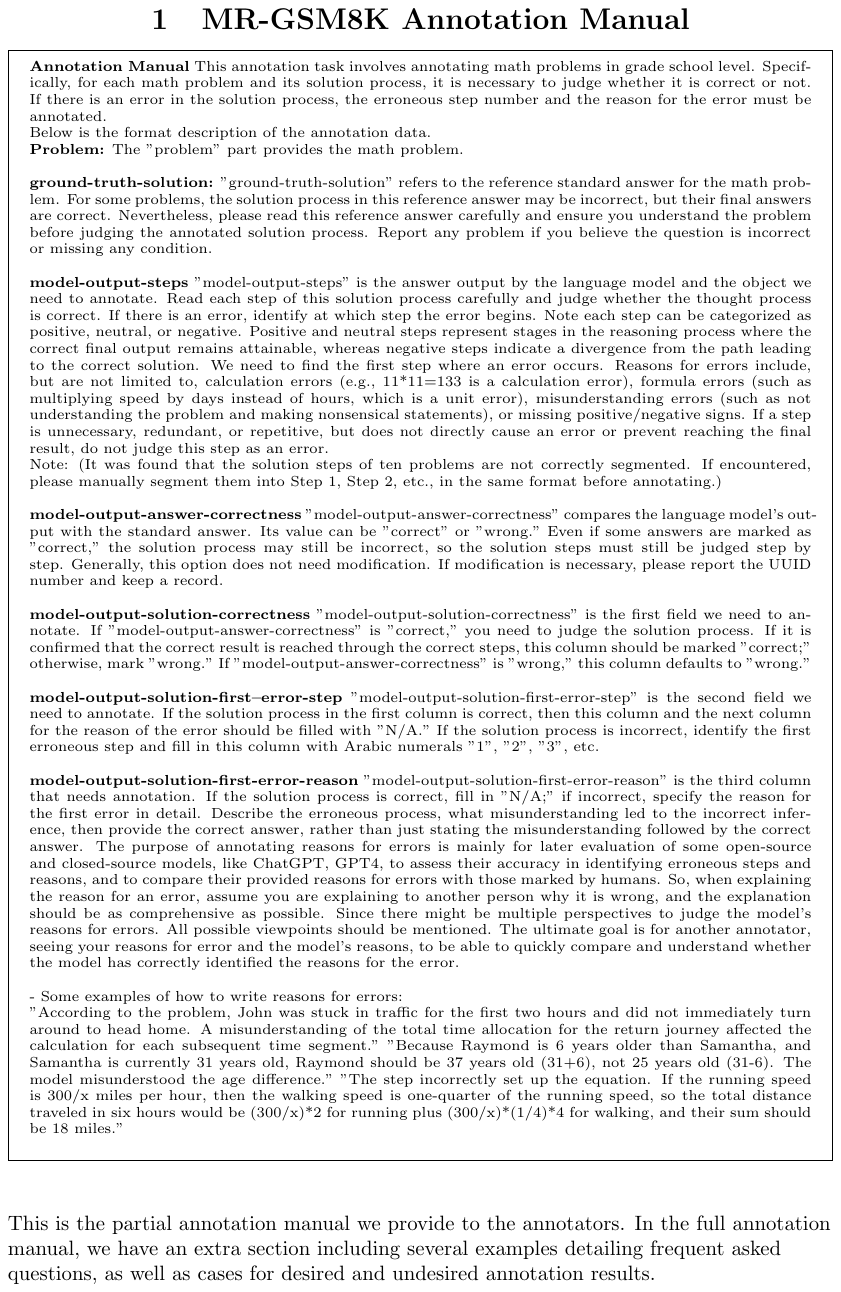}

\end{document}